\pdfoutput=1

\documentclass[11pt]{article}

\usepackage{acl}
\usepackage{times}
\usepackage{latexsym}

\usepackage[T1]{fontenc}

\usepackage[utf8]{inputenc}

\usepackage{times}
\usepackage{latexsym}
\usepackage{algorithm, algorithmic}
\usepackage{amsmath} 
\usepackage{amssymb}
\usepackage{multirow}
\usepackage{makecell}
\usepackage{hhline}
\usepackage{xspace}
\usepackage{booktabs}
\usepackage{float}
\usepackage{graphicx}

\usepackage{amssymb}
\usepackage{pifont}
\usepackage{bm}
\usepackage{enumitem}
\usepackage{diagbox}
\setlist[itemize]{leftmargin=*}
\setitemize[1]{itemsep=0pt,partopsep=0pt,parsep=\parskip,topsep=0pt}
\usepackage{microtype}

\title{Graph Enhanced Contrastive Learning for Radiology Findings Summarization}

\author{%
Jinpeng Hu$^{\spadesuit\heartsuit *}$, \hspace{0.2cm}
Zhuo Li$^{\spadesuit\heartsuit *}$, \hspace{0.2cm}
Zhihong Chen$^{\spadesuit\heartsuit}$, \hspace{0.2cm}
Zhen Li$^{\spadesuit\heartsuit}$ \\
 \textbf{Xiang Wan}$^{\heartsuit\diamondsuit\dag}$, \hspace{0.2cm} \textbf{Tsung-Hui Chang}$^{\spadesuit\heartsuit\dag}$ \\
$^{\spadesuit}$The Chinese University of Hong Kong (Shenzhen) \\ 
$^{\heartsuit}$Shenzhen Research Institute of Big Data \hspace{0.2cm} \\
$^{\diamondsuit}$Pazhou Lab, Guangzhou, 510330, China \hspace{0.2cm} \\
\texttt{
$^\spadesuit$\{jinpenghu, zhuoli3, zhihongchen\}@link.cuhk.edu.cn} \\
\texttt{$^{\heartsuit}$wanxiang@sribd.cn} \hspace{0.2cm}
\texttt{$^{\spadesuit}$\{lizhen, changtsunghui\}@cuhk.edu.cn}
}

\begin{document}

\maketitle

\def\thefootnote{*}\footnotetext{Equal Contribution.}
\def\thefootnote{\dag}\footnotetext{Corresponding author.}
\renewcommand{\thefootnote}{\arabic{footnote}}

\begin{abstract}
The impression section of a radiology report summarizes the most prominent observation from the findings section and is the most important section for radiologists to communicate to physicians.
Summarizing findings is time-consuming and can be prone to error for inexperienced radiologists, and thus automatic impression generation has attracted substantial attention.
With the encoder-decoder framework, most previous studies explore incorporating extra knowledge (e.g., static pre-defined clinical ontologies or extra background information).
Yet, they encode such knowledge by a separate encoder to treat it as an extra input to their models, which is limited in leveraging their relations with the original findings.
To address the limitation, we propose a unified framework for exploiting both extra knowledge and the original findings in an integrated way so that the critical information (i.e., key words and their relations) can be extracted in an appropriate way to facilitate impression generation.
In detail, for each input findings, it is encoded by a text encoder, and a graph is constructed through its entities and dependency tree.
Then, a graph encoder (e.g., graph neural networks (GNNs)) is adopted to model relation information in the constructed graph.
Finally, to emphasize the key words in the findings, contrastive learning is introduced to map positive samples (constructed by masking non-key words) closer and push apart negative ones (constructed by masking key words).
The experimental results on OpenI and MIMIC-CXR confirm the effectiveness of our proposed method.\footnote{Our code is released at \url{https://github.com/jinpeng01/AIG_CL}.}
\end{abstract}

\section{Introduction}
Radiology reports document critical observation in a radiology study and play a vital role in communication between radiologists and physicians.
A radiology report usually consists of a findings section describing the details of medical observation and an impression section summarizing the most prominent observation.
The impression is the most critical part of a radiology report, but the process of summarizing findings is normally time-consuming and could be prone to errors for inexperienced radiologists.
Therefore, automatic impression generation (AIG) has drawn substantial attention in recent years, and there are many methods proposed in this area \cite{zhang2018learning,attend,macavaney2019ontology,shieh2019towards}.

\begin{figure}[t]
\centering
\includegraphics[width=0.48\textwidth, trim=0 0 -20 0]{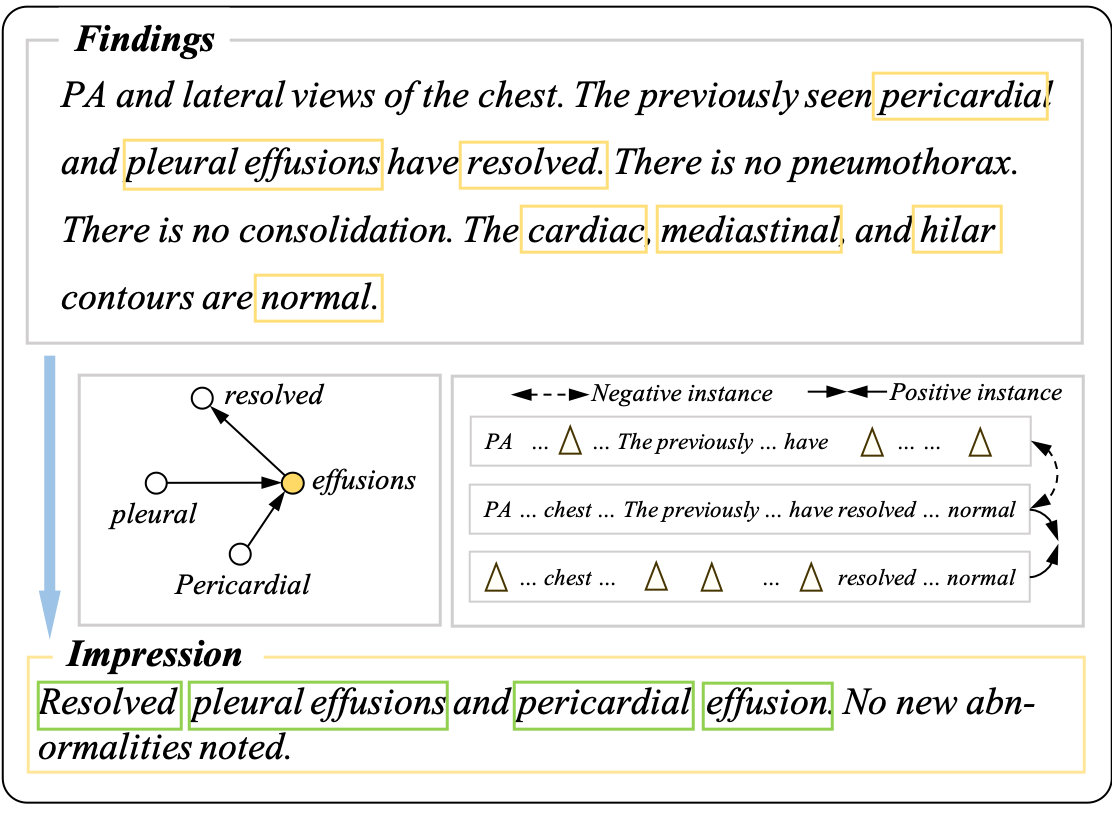}
\caption{An example of the findings and corresponding impression, where the relation information, as well as positive and negative examples, are also shown in the figure. Note that $\triangle$ represents the removed word.}
\label{fig:example_1}
\vskip -1em
\end{figure}

\begin{figure*}[t]
\centering
\includegraphics[width=0.98\textwidth]{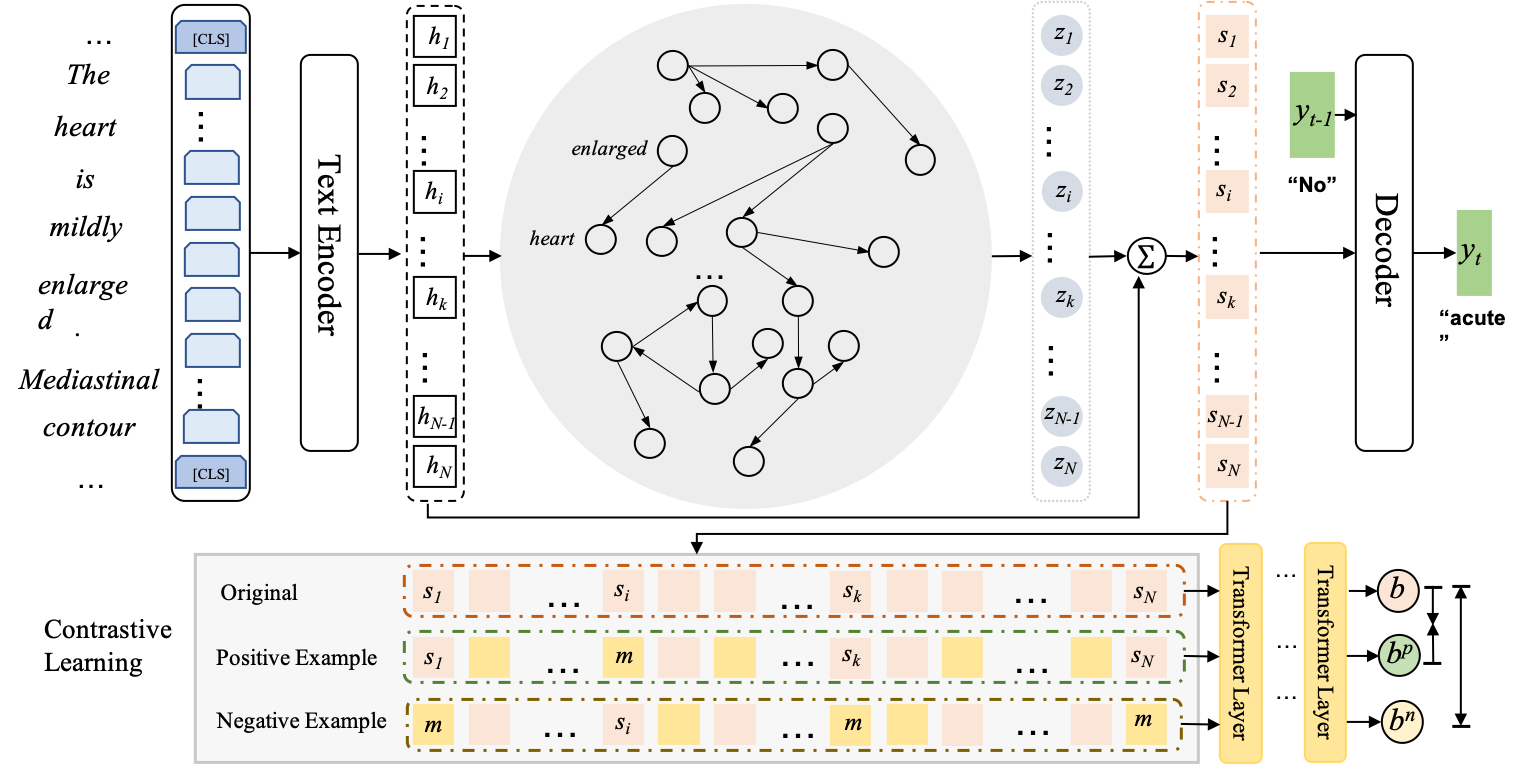}
\vskip -1em
\caption{The overall architecture of our proposed method with graph and contrastive learning. An example input and output at $t-1$ and $t$ step are shown in the figure, where the top is the backbone sequence-to-sequence paradigm with a graph to store relation information between critical words and the bottom is the contrastive learning module with specific positive and negative examples. $\mathbf{m}$ refer to a mask vector.}
\label{fig:architecture}
\vskip -1em
\end{figure*}

Most existing studies focus on incorporating extra knowledge on the general encoder-decoder framework.
For example, \newcite{zhang2018learning} utilized the background section in the radiology report through a separate encoder and then used it to guide the decoding process to enhance impression generation.
Similarly, \newcite{macavaney2019ontology} and \newcite{attend} proposed to extract the ontology information from findings and used an encoder to encode such information to promote the decoding process.
Although these approaches have brought significant improvements, they only leverage extra knowledge and findings separately (i.e., through an extra encoder).
Thus, their performance relies heavily on the quality of extra knowledge, and the further relationships between extra knowledge and findings are not explored.
In this paper, we propose a unified framework to exploit both findings and extra knowledge in an integrated way so that the critical information (i.e., key words and their relations in our paper) can be leveraged in an appropriate way.
In detail, for each input findings, we construct a word graph through the automatically extracted entities and dependency tree, with its embeddings, which are from a text encoder.
Then, we model the relation information among key words through a graph encoder (e.g., graph neural networks (GNNs)).
Finally, contrastive learning is introduced to emphasize key words in findings to map positive samples (constructed by masking non-key words) closer and push apart negative ones (constructed by masking key words), as shown in Figure \ref{fig:example_1}.
In such a way, key words and their relations are leveraged in an integrated way through the above two modules (i.e., contrastive learning and the graph encoder) to promote AIG.
Experimental results on two datasets (i.e., OpenI and MIMIC-CXR) show that our proposed approach achieves state-of-the-art results.
\section{Method}
\textcolor{black}{
We follow the standard sequence-to-sequence paradigm for AIG. First, we utilize WordPiece \cite{wu2016google} to tokenize original findings and obtain the source input sequence $\mathcal{X} = x_{1},x_{2},\cdots, x_{N}$, where $N$ is the number of tokens in $\mathcal{X}$.
The goal is to find a sequence $\mathbf{Y}=\left\{y_{1},...y_{i},..., y_{L}\right\}$ that summarizes the most critical observations in findings, where $L$ is the length of impression and $y_{i} \in \mathrm{V}$ are the generated tokens and $\mathrm{V}$ is the vocabulary of all possible tokens.
The generation process can be formalized as:
\begin{equation}
\setlength\abovedisplayskip{6pt}
\setlength\belowdisplayskip{6pt}
    p(\mathbf{Y} \mid \mathcal{X})=\prod_{t=1}^{L} p\left(y_{t} \mid y_{1}, \ldots, y_{t-1}, \mathcal{X}\right)    
\end{equation}
The model is then trained to maximize the negative conditional log-likelihood of $\mathcal{Y}$ given the $\mathcal{X}$:
\begin{equation}
\setlength\abovedisplayskip{6pt}
\setlength\belowdisplayskip{6pt}
    \theta^{*}=\underset{\theta}{\arg \max}  \sum_{t=1}^{L} \log p\left(y_{t} \mid y_{1}, ...,  y_{t-1}, \mathcal{X}, A;\theta\right)
\end{equation}
where $\theta$ is the parameters of the model, and $A$ represents edges in the relation graph.
An overview of our proposed method is presented in Figure \ref{fig:architecture}.
Our model contains three main components, i.e., the graph enhanced encoder, the contrastive learning module, and the decoder. The details are described in the following sub-sections.
}
\subsection{Relation Graph}
\label{relation_graph}
\textcolor{black}{
The impression usually describes critical abnormalities with more concise descriptions summarized from the corresponding findings and sometimes uses key phrases to express observations.
For example, a sentence in findings texts \textit{``There is a left pleural effusion which is small in size.''}, is simplified as a key phrase \textit{``Small left pleural effusion''} in the impression, where the relation between \textit{``small''} and \textit{``effusion''} is vital for describing the corresponding observation.
Thus, the relation information in findings plays an essential role in accurate key phrase generation.
Four types of medical entities, \textit{anatomy}, \textit{observation}, \textit{anatomy modifier}, and \textit{observation modifier}, are recognized from findings, which compose a majority of important medical knowledge in impression \cite{hassanpour2016information}.
With WordPiece tokenization, we represent each entity by frequent subwords and connect any two subwords if they are adjacent in the same entity to enhance internal relations for keeping the entity complete.
For example, the entity \textit{``opacity''} is represented as \textit{``op \#\#acity''} and then these two subwords connect to each other with both from \textit{``op''} to \textit{``\#\#acity''} and from \textit{``\#\#acity''} to \textit{``op''}.
Besides, we need to consider the semantic relation between entities and other words, such as words used to describe the location and degree of symptoms, which is necessary for accurately recording abnormalities.
For example, in a text span \textit{``bilateral small pleural effusions''}, relations in <\textit{``bilateral''},\textit{``effusions''}>, <\textit{``small''},\textit{``effusions''}> are also important to describe the observation \textit{``effusions''} and they can be extracted from the dependency tree.
Therefore, we construct a dependency tree to extract the semantic relations between entities and other words, with the direction from their head words to themselves.
We also employ the WordPiece to split these words as subwords and connect all the source subwords to the corresponding target words with the original direction.
The constructed subword graph is then used to extract relation information, with edges represented by $A$.
}
\begin{table}[!]
\centering
\begin{tabular}{l}
\toprule[1pt]
\textbf{Algorithm 1:} Generation of Examples \\ \hline
\textbf{Input:} $\mathbf{s}$: graph enhanced token representation \\ 
\quad \quad \quad $A$: edges in relation graph \\
\textbf{Output:} $\mathbf{s}^{p}$ Positive example \\
\quad \quad \quad \ \ \ $\mathbf{s}^{n}$ Negative example \\
\textbf{Initialization:} $\mathbf{s}^{p} \leftarrow \mathbf{s}$,  $\mathbf{s}^{n} \leftarrow \mathbf{s}$\\
\quad \quad \quad \quad \quad \ \ \ \ $\mathbf{m}=[1e-6] \in \mathrm{R}^{d}$\\
1: $N$, $d$ = \emph{size}($\mathbf{s}$) \\
2: $V_{key}$ = \emph{Extract\_subword\_index}($A$) \\
3: \textbf{for} $j=0$ to $N$ \textbf{do} \\
4: \quad \quad \textbf{if} j in $V_{key}$ \textbf{then}\\
5: \quad \quad \quad \quad $\mathbf{s}^{n}_{j} \leftarrow \mathbf{m}$ \\
6: \quad \quad \textbf{else}: \\
7: \quad \quad \quad \quad $\mathbf{s}^{p}_{j} \leftarrow \mathbf{m}$ \\
8: \quad \quad \textbf{end if} \\
9: \textbf{end for} \\
\bottomrule
\end{tabular}
\label{Tab:al}
\vskip -1em
\end{table}

\subsection{Graph Enhanced Encoder}
\textcolor{black}{
In recent years, pre-trained models have dominated not only general summarization tasks but also multi-modal tasks because of their strong ability in feature representation \cite{wu-2021-bass,zhang2020pegasus, yuan2021instancerefer, yuan2022x}.
Thus, in our method, we utilize the pre-trained model BioBERT \cite{BioBERT} trained on a large biomedical corpus as our text encoder.
The hidden state $\mathbf{h}_{i}$ for each token $x_{i}$ is generated by the text encoder
\begin{equation}
\setlength\abovedisplayskip{5pt}
\setlength\belowdisplayskip{5pt}
    [\mathbf{h}_{1},\mathbf{h}_{2},\cdots,\mathbf{h}_{n}] = {f}_{te}(x_1,x_2,\cdots ,x_n)
\end{equation}
Herein, ${f}_{te}(\cdot)$ refers to the pre-trained Transformer-based text encoder (i.e., BioBERT \cite{BioBERT}), and $\mathbf{h}_{i}$ is a $d$-dimensional feature vector for representing corresponding tokens $x_{i}$.
Since GNNs are well known for extracting features from graph structure and have been shown promising in text generation tasks \cite{jia2020neural,hu2021word}, we employ a GNN-based encoder to capture relation information from the corresponding subword graph.
This process can be formulated as:
\begin{equation}
\setlength\abovedisplayskip{6pt}
\setlength\belowdisplayskip{6pt}
    \mathbf{z} = f_{ge}(\mathbf{h},A),
    \label{eq:graph-encoder}
\end{equation}
where $f_{ge}(\cdot)$ is the graph encoder, and $\mathbf{z}$ is the feature vector extracted from the graph.
Next, to incorporate relation information into token representation, we concatenate $\mathbf{z}$ and $\mathbf{h}$ and utilize a fully connected layer to reduce it to the same dimensions as $\mathbf{z}$ and $\mathbf{h}$:
\begin{equation}
\setlength\abovedisplayskip{6pt}
\setlength\belowdisplayskip{6pt}
    \mathbf{s} = \text{MLP}([\mathbf{h}_{1}\oplus\mathbf{z}_{1},\mathbf{h}_{2}\oplus\mathbf{z}_{2},\cdots,\mathbf{h}_{n}\oplus\mathbf{z}_{n}]),
    \label{eq:graph-encoder}
\end{equation}
where $\mathbf{s}$ is the final token representation.
}
\subsection{Contrastive Learning}
\textcolor{black}{
Only relying on a GNN encoder to capture relation information still lacks the capability to fully grasp important word information from findings since the graph is pre-defined before training or testing.
Recently, contrastive learning has shown strong power in learning and distinguishing significant knowledge by concentrating positive samples and contrasting with negative samples, and brought significant improvements in many tasks, such as improving the faithfulness of summarization and discriminating vital information to enhance representation \cite{cao2021cliff,zeng2021modeling}.
We expect our model to be more sensitive to critical words contained in findings.
For this purpose, we apply a contrastive learning module to concentrate positive pairs and push negative ones apart, which aims to help the model differentiate essential information from secondary information.
We regard tokens with edges in the relation graph as critical tokens since they contain important information for describing key observations, as discussed in \ref{relation_graph}.
To construct a positive example, we mask each non-key token representation in $\mathbf{s}$ as the constant vectors $\mathbf{m} \in \mathbb{R}^{d}$, with all elements $1e-6$, so that this instance can consolidate the critical information and remove unimportant words.
Meanwhile, we utilize a similar way to mask important token representations in $\mathbf{s}$ as $\mathbf{m}$ to obtain a negative example $\mathbf{s}^{n}$.
The details of generating positive and negative examples are shown in Algorithm 1.
Note that in our model, we do not consider the other instances in the same mini-batch as the negative examples, which is different from many existing approaches \cite{kim2021self,giorgi2020declutr} since we aim to identify the critical content in $\mathcal{X}$ instead of expanding differences between various findings in one mini-batch.
In addition, radiology reports are not as diverse as ordinary texts, and they are mainly composed of fixed medical terms and some attributive words, where the former is used to record critical information and the latter is to keep sentences fluent and grammatically correct.
}

\textcolor{black}{
Afterward, we employ a randomly initialized Transformer-based encoder to model $\mathbf{s}$, $\mathbf{s}^{p}$, $\mathbf{s}^{n}$, respectively, which can be formulated as:
\begin{align}
\setlength\abovedisplayskip{6pt}
\setlength\belowdisplayskip{6pt}
    \mathbf{b}  &= f_{ce}(\mathbf{s}), \\
    \mathbf{b}^{p} &= f_{ce}(\mathbf{s}^{p}), \\
    \mathbf{b}^{n} &= f_{ce}(\mathbf{s}^{n}),
    \label{eq:cl-encoder}
\end{align}
where $f_{ce}(\cdot)$ represents the contrastive encoder.
$\mathbf{b}$, $\mathbf{b}^{p}$ and $\mathbf{b}^{n}$ are intermediate states extracted from the encoder, which are also $d$-dimensional vectors. 
Then, we calculate cosine similarity $sim(\mathbf{b}^{1},\mathbf{b}^{2}) = \frac{\mathbf{b}_{1}^{\top} \mathbf{b}_{2}}{\left\|\mathbf{b}_{1}\right\| \cdot\left\|\mathbf{b}_{2}\right\|}$ for positive and negative pairs, denoted as $sim(\mathbf{b},\mathbf{b}^{p})$ and $sim(\mathbf{b},\mathbf{b}^{n})$.
We follow \newcite{robinson2020contrastive} to formulate the training objective of contrastive module:
\begin{equation}
\setlength\abovedisplayskip{5pt}
\setlength\belowdisplayskip{5pt}
    l_{con} = -\log \frac{e^{\operatorname{sim}\left(\mathbf{b}_{i}, \mathbf{b}^{p}\right) / \tau}}{\sum_{j=1}^{N}\left(e^{\operatorname{sim}\left(\mathbf{b}_{i}, \mathbf{b}^{p}\right) / \tau}+e^{\operatorname{sim}\left(\mathbf{b}_{i}, \mathbf{b}^{n}\right) / \tau}\right)},
\end{equation}
where $\tau$ is a temperature hyperparameter, which is set to 1 in this paper.
}

\subsection{Decoder}
\textcolor{black}{
The decoder in our model is built upon a standard Transformer \cite{vaswani2017attention}, where the representation $\mathbf{s}$ is functionalized as the input of the decoder so as to improve the generation process.
In detail, $\mathbf{s}$ is sent to the decoder at each decoding step, jointly with the generated tokens from previous steps, and thus the current output $y_{t}$ can be computed by
\begin{equation}
\setlength\abovedisplayskip{5pt}
\setlength\belowdisplayskip{5pt}
    y_{t} = f_e(\mathbf{s}_{1},\mathbf{s}_{2},\cdots,\mathbf{s}_{n},y_{1},\cdots,y_{t-1}),
\end{equation}
where $f_e(\cdot)$ refers to the Transformer-based decoder and this process is repeated until the complete impression is obtained.
}

\textcolor{black}{
Besides, to effectively incorporate the critical word information into the decoding process, we sum the losses from the impression generation and contrastive objectives as
\begin{equation}
\setlength\abovedisplayskip{5pt}
\setlength\belowdisplayskip{5pt}
    L = l_{ge}+ \lambda l_{con},
\end{equation}
where $l_{ge}$ is the basic sequence-to-sequence loss, and $\lambda$ is the weight to control the contrastive loss.
}

\begin{table}[t]
\footnotesize
\centering
\resizebox{.46\textwidth}{!}{
\begin{tabular}{l|l|r|r|r}
\toprule[1pt]
{\textsc{\textbf{Data}}}&\textsc{\textbf{Type}}  &\textsc{\textbf{Train}} & \textsc{\textbf{Dev}} &\textsc{\textbf{Test}} \\
\midrule                       
\multirow{5}{*} {\makecell*[l]{\textsc{OpenI}}}
& \textsc{Report \#} & {2400} & {292} & {576}\\
& \textsc{Avg. wf} & {37.89} & {37.77}  & {37.98} \\
& \textsc{Avg. sf} & {5.75} & {5.68}  & {5.77} \\
& \textsc{Avg. wi} & {10.43} & {11.22}  & {10.61} \\
& \textsc{Avg. si} & {2.86} & {2.94}  & {2.82} \\
\midrule
\multirow{5}{*} {\makecell*[l]{\textsc{MIMIC} \\ {-CXR}}}
& \textsc{Report \#} & {122,014} & {957} & {1,606}\\
& \textsc{Avg. wf} & {55.78} & {56.57}  & {70.67} \\
& \textsc{Avg. sf} & {6.50} & {6.51}  & {7.28} \\
& \textsc{Avg. wi} & {16.98} & {17.18}  & {21.71} \\
& \textsc{Avg. si} & {3.02} & {3.04}  & {3.49} \\
\bottomrule
 \end{tabular}}
\vskip -0.3em
\linespread{1}
  \caption{The statistics of the two benchmark datasets with random split for \textsc{OpenI} and official split for \textsc{MIMIC-CXR}, including the numbers of report, the averaged sentence-based length (\textsc{Avg. sf}, \textsc{Avg. si}), the averaged word-based length (\textsc{Avg. wf}, \textsc{Avg. wi}) of both \textsc{Impression} and \textsc{Findings}.}%
  \label{Tab:dataset}
\vskip -1em
\end{table}
\begin{table*}[t]
\footnotesize
\centering
\resizebox{.94\textwidth}{!}{
\begin{tabular}{l|p{3.5cm}|ccc|ccc}
\toprule[1pt]
\multirow{2}{*}{\textsc{\textbf{\makecell[c]{Data}}}} & \multirow{2}{*}{\textsc{\textbf{\makecell[c]{Model}}}} 
& \multicolumn{3}{c|}{\textsc{\textbf{ROUGE}}} & \multicolumn{3}{c}{\textsc{\textbf{FC}}}  \\  
& & \textsc{R-1}  & \textsc{R-2}  & \textsc{R-L}   &\textsc{\textbf{P}} & \textsc{\textbf{R}} &\textsc{\textbf{F-1}} \\
\midrule                       
\multirow{4}{*} {\makecell*[l]{\textsc{OpenI}}}

& \textsc{Base} & {62.74} & {53.32} & {62.86} & {-} & {-} & {-}\\
& \textsc{Base+CL} & {63.53} & {54.58}  & {63.13} & {-} & {-} & {-}\\
& \textsc{Base+graph} & {63.29} & {54.12}  & {63.03} & {-} & {-} & {-}\\

& \textsc{Base+graph+CL} & \textbf{{64.97}} & \textbf{{55.59}}  & \textbf{{64.45}} & {-} & {-} & {-}\\
\midrule
\multirow{4}{*} {\makecell*[l]{\textsc{MIMIC-CXR}}}
& \textsc{Base} & {47.92} & {32.43} & {45.83} & {58.05} & {50.90} & {53.01} \\

& \textsc{Base+CL} & {48.15} & {33.25}  & {46.24} & {58.34} & {51.58} & {53.70} \\

& \textsc{Base+graph} & {48.29} & {33.30}  & {46.36} & {57.80} & {51.70} & {53.50} \\

& \textsc{Base+graph+CL} & \textbf{{49.13}} & \textbf{{33.76}}  & \textbf{{47.12}}   
& \textbf{58.85} & \textbf{52.33}  & \textbf{54.52} \\
\bottomrule
 \end{tabular}
 }
\vskip -2mm
  \caption{Comparisons of baselines and our method on \textsc{OpenI} and \textsc{MIMIC-CXR} datasets. R-1, R-2 and R-L refer to ROUGE-1, ROUGE-2 and ROUGE-L, respectively. \textsc{P}, \textsc{R} and \textsc{F-1} represent precision, recall, and F1 score.}%
  \label{Tab:performance_on_different_base}
\vskip -4mm
\end{table*}

\section{Experimental Setting}
\subsection{Dataset}
\textcolor{black}{
Our experiments are conducted on two following datasets: \textbf{\textsc{OpenI}} \cite{demner2016preparing} and \textbf{\textsc{MIMIC-CXR}} \cite{johnson2019mimic} respectively, where the former contains 3268 reports collected by Indiana University and the latter is a larger dataset containing 124577 reports.
Note that the number of reports we introduced is counted after pre-processing.
We follow \cite{hu2021word,zhang2018learning} to filter the reports by deleting the reports in the following cases: (1) no findings or no impression sections; (2) the findings have fewer than ten words, or the impression has fewer than two words.
For \textsc{OpenI}, we follow \cite{hu2021word} to randomly divide it into train/validation/test set by 2400:292:576 in our experiments.
For \textsc{MIMIC-CXR}, we apply two types of splits, including an official split and a random split with a ratio of 8:1:1 similar to \cite{attend}.
We report the statistics of these two datasets in Table \ref{Tab:dataset}.
}
\begin{table*}[ht]
\linespread{1.15}
\small
\centering
\begin{tabular}{l|rrr|rrr|rrr}
\toprule
{\multirow{3}{*}
{ \textsc{\textbf{\makecell[l]{\\Model}}}}}  & \multicolumn{3}{c|}{\textsc{\textbf{OpenI}}} & \multicolumn{6}{c}{\textsc{\textbf{MIMIC-CXR}}} \\
\cmidrule(r){2-10}
& \multicolumn{3}{c|}{\textsc{\textbf{Random split}}} & \multicolumn{3}{c|}{\textsc{\textbf{Official split}}}   & \multicolumn{3}{c}{\textsc{\textbf{Random split}}} \\  
                                                                                   & \textsc{R-1}  & \textsc{R-2}  & \textsc{R-L}   & \textsc{R-1}        & \textsc{R-2}       & \multicolumn{1}{c|}{\textsc{R-L}}         & \textsc{R-1}        & \textsc{R-2}       & \textsc{R-L}      \\
\cmidrule(lr){1-10}
\textsc{LexRank} \cite{erkan2004lexrank}   &14.63    &4.42    &14.06    &18.11    & 7.47    & 16.87      & -      & -       & -  \\

\textsc{TransExt} \cite{liu2019text} & 15.58    & 5.28     & 14.42    & 31.00   &16.55   &27.49   & -   & -   & -      \\
\textsc{PGN (LSTM)} \cite{see2017get}  & 63.71   & 54.23    & 63.38    &46.41    & 32.33     & 44.76    & -    & -     & -     \\
\textsc{TransAbs} \cite{liu2019text}  & 59.66   & 49.41    & 59.18    & 47.16    & 32.31     & 45.47    & -    & -     & -     \\

$\textsc{OntologyABS}^{\dagger}$ \cite{attend} & -  & -   & -   & -    & -    & -  & 53.57   &40.78  & 51.81 \\
$\textsc{WGSum (LSTM)}^{\dagger}$ \cite{hu2021word} &{{64.32}} & {{55.48}}  & {{63.97}} &{47.48} &{33.03} &{45.43} & {54.97}  & {43.64}  & {53.81} \\

$\textsc{WGSum (Trans)}^{\dagger}$ \cite{hu2021word} &{61.63} & {50.98}  & {61.73}   & {{48.37}} & {{33.34}}  & {{46.68}}       & {56.38}    & {44.75}   & {55.32}  \\
\cmidrule(lr){1-10}
\textsc{Ours}& \textbf{{64.97}} & \textbf{{55.59}}  & \textbf{{64.45}}
& \textbf{{49.13}} & \textbf{{33.76}}  & \textbf{{47.12}}   
& \textbf{57.38}    & \textbf{45.52}   & \textbf{56.13}  \\
\bottomrule
\end{tabular}
\linespread{1}
\caption{Comparisons of our proposed models with previous study on the \textsc{OpenI} and \textsc{MIMIC-CXR} with respect to ROUGE metric. $\dagger$ refers to that the results is directly cited from the original paper.}
\label{Tab:performance_on_different_model}
\vskip -1em
\end{table*}

\subsection{Baseline and Evaluation Metrics}
\textcolor{black}{
To explore the performance of our method, we use the following ones as our main baselines:
\vskip -1em
\begin{itemize}[leftmargin=*]
    \setlength{\topsep}{0pt}
    \setlength{\itemsep}{0pt}
    \setlength{\parsep}{0pt}
    \setlength{\parskip}{0pt}
    \item \textbf{\textsc{Base}} \cite{liu2019text}: this is a backbone sequence-to-sequence model, i.e., a pre-trained encoder and a randomly initialized Transformer-based decoder.
    \item \textbf{\textsc{Base+graph}} and \textbf{\textsc{Base+CL}}: these have the same architecture as \textbf{\textsc{Base}}, where the former incorporates an extra graph encoder to enhance relation information, and the latter introduces a contrastive learning module to help the model distinguish critical words.
\end{itemize}
}
\noindent
\textcolor{black}{
Besides, we also compare our method with those existing studies, including both extractive summarization methods, e.g., \textbf{\textsc{LexRank}} \cite{erkan2004lexrank}, \textbf{\textsc{TransformerEXT}} \cite{liu2019text}, and the ones proposed for abstractive models. e.g., \textbf{\textsc{TransformerABS}} \cite{liu2019text}, \textbf{\textsc{OntologyABS}} \cite{attend}, \textbf{\textsc{WGSum (Trans+GAT)}}, and \textbf{\textsc{WGSum (LSTM+GAT)}} \cite{hu2021word}.
}

Actually, factual consistency (FC) is critical in radiology report generation \cite{liu2019clinically,chen2020generating}.
Following \newcite{zhang2020optimizing,hu2021word}, we evaluate our model and three baselines by two types of metrics: summarization and FC metrics.
For summarization metrics, we report $\textsc{F}_{1}$ scores of ROUGE-1 (R-1), ROUGE-2 (R-2), and ROUGE-L (R-L). 
Besides, for FC metrics, we utilize CheXbert \cite{smit2020chexbert}\footnote{FC is only applied to MIMIC-CXR since the CheXbert is designed for MIMIC-CXR. We obtain it from \url{https://github.com/stanfordmlgroup/CheXbert}} to detect 14 observations related to diseases from reference impressions and generated impressions and then calculate the precision, recall, and F1 score between these two identified results.

\subsection{Implementation Details}
In our experiments, we utilize biobert-base-cased-v1.1\footnote{We obtain BioBERT from \url{https://github.com/dmis-lab/biobert}} as our text encoder and follow its default model settings: we use 12 layers of self-attention with 768-dimensional embeddings.
Besides, we employ stanza \cite{zhang2020biomedical} to extract medical entities and the dependence tree, which is used to construct the graph and generate positive and negative examples.
Our method is implemented based on the code of BertSum \cite{liu2019text}\footnote{We obtain the code of BertSum from \url{https://github.com/nlpyang/PreSumm}}.
In addition, we use a 2-layer graph attention networks (GAT) \cite{velivckovic2017graph}\footnote{Since previous study \cite{hu2021word} has shown that GAT \cite{velivckovic2017graph} is more effective in impression generation, we select GAT as our graph encoder.} with the hidden size of 768 as our graph encoder and a 6-layer Transformer with 768 hidden sizes and 2048 feed-forward filter sizes for the contrastive encoder.
The decoder is also a 6-layer Transformer with 768 dimensions, 8 attention heads, and 2048 feed-forward filter sizes.
Note that $\lambda$ is set 1 in all experiments, and more detailed hyperparameters are reported in \ref{appendix:hyperparameter}.
During the training, we use Adam \cite{kingma2014adam} to optimize the trainable parameters in our model.

\section{Results and Analyses}

\subsection{Effect of Graph and Contrastive learning}

To explore the effectiveness of our proposed method, we conduct experiments on two benchmark datasets, with the results reported in Table \ref{Tab:performance_on_different_base}, where \textsc{Base+graph+CL} represents our complete model.
We can obtain several observations from the results.
First, both \textsc{Base+graph} and \textsc{Base+CL} achieve better results than \textsc{Base} with respect to R-1, R-2, and R-L, which indicates that graph and contrastive learning can respectively promote impression generation.
Second, \textsc{Base+graph+CL} outperforms all baselines with significant improvement on two datasets, confirming the effectiveness of our proposed method in combining graph and contrastive learning.
This might be because graphs and contrastive learning can provide valuable information from different aspects, the former mainly record relation information, and the latter brings critical words knowledge, so that an elaborate combination of them can bring more improvements.
Third, when comparing these two datasets, the performance gains from our full model over three baselines on OpenI are more prominent than that on MIMIC-CXR.
This is perhaps because compared to MIMIC-CXR, OpenI dataset is relatively smaller and has a shorter averaged word-based length, such that it is easier for the graph to record relation and more accessible for contrastive learning to recognize key words by comparing positive and negative examples.
Fourth, we can find a similar trend on the FC metric on the MIMIC-CXR dataset, where a higher F1 score means that our complete model can generate more accurate impressions thanks to its more substantial power in key words discrimination and relationship information extraction.

\subsection{Comparison with Previous Studies}
\textcolor{black}{
In this subsection, we further compare our models with existing models on the aforementioned datasets, and the results are reported in Table \ref{Tab:performance_on_different_model}.
There are several observations.
First, the comparison between our model and \textsc{OntologyABS} shows the effectiveness of our design in this task, where our model achieves better performance, though both of them enhance impression generation by incorporating crucial medical information.
This might be because by comparing positive and negative examples for each findings, our model is more sensitive to critical information and more intelligent in distinguishing between essential information and secondary information, contributing to more accurate and valuable information embedded in the model. 
Second, we can observe that our model outperforms all existing models in terms of R-1, R-2, and R-L.
On the one hand, effectively combining contrastive learning and graph into the sequence to sequence model is a better solution to improve feature extraction and thus promote the decoding process robustly.
On the other hand, the pre-trained model (i.e., BioBERT) used in our model is a more powerful feature extractor in modeling biomedical text than those existing studies, e.g., \textsc{TransformerABS}, \textsc{OntologyABS}, and \textsc{PGN}, which utilize randomly initialized encoders.
Third, when compared to those complicated models, e.g., WGSUM utilize stanza to extract entities and construct two extra graph encoders to extract features from a word graph, which are then regarded as background information and dynamic guiding information to enhance the decoding process for improving impression generation, our model can achieve better performance through a somewhat more straightforward method.
}

\begin{figure}[t]
\centering
\includegraphics[width=0.5\textwidth, trim=0 0 0 10]{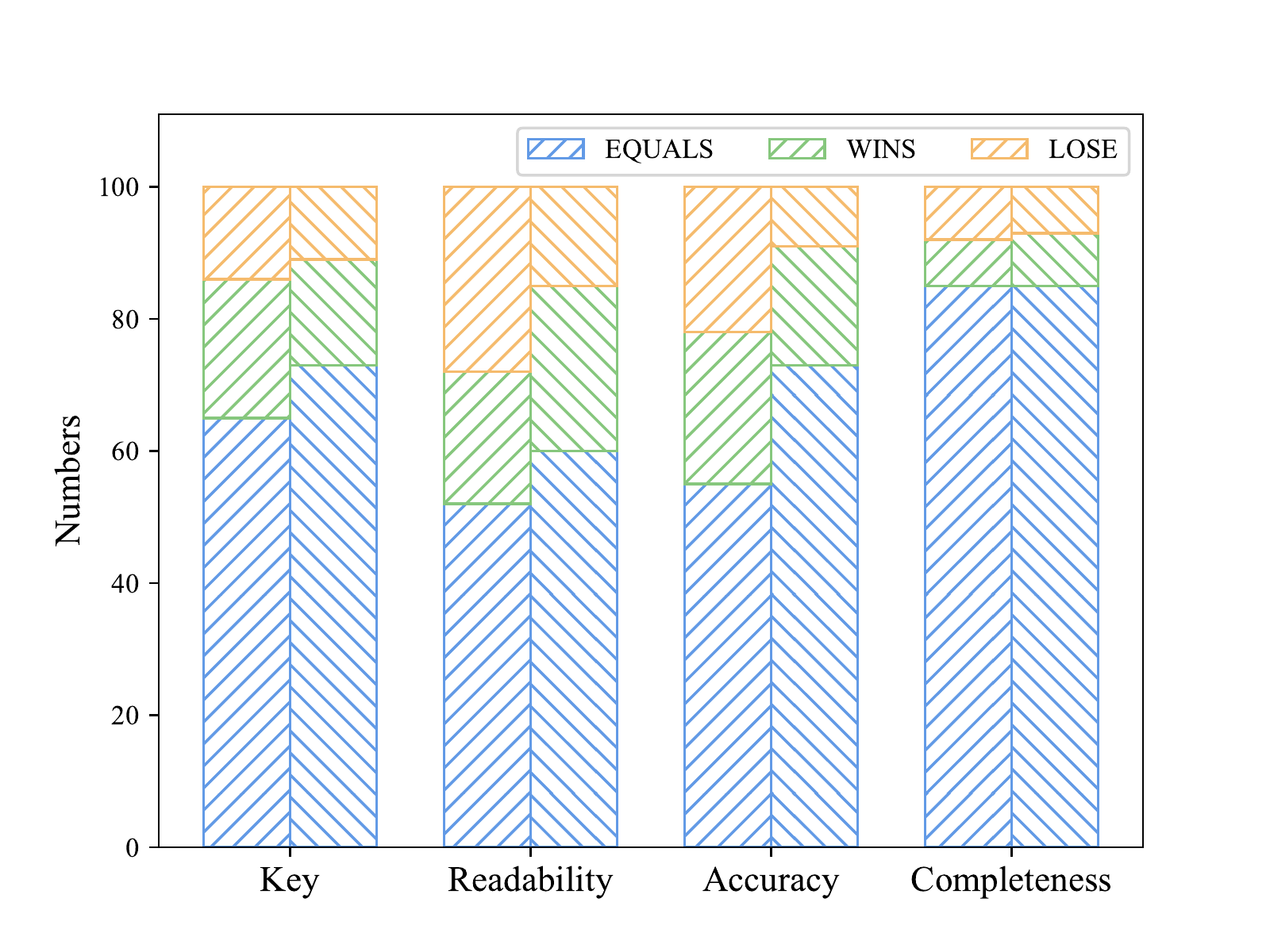}
\vskip -0.5em
\caption{The results of human evaluation, where forward and backslash represent that \textsc{Base+graph+CL} versus the reference and \textsc{Base}, respectively. Yellow, green and blue represent that our model loses, equal to competitors and wins.}
\label{fig:human_evaluation}
\vskip -1em
\end{figure}

\subsection{Human Evaluation}
\textcolor{black}{
We further conduct a human evaluation to understand the quality of the generated impression better and alleviate the limitation of the ROUGE metric.
One hundred generated impressions on MIMIC-CXR from \textsc{Base} and \textsc{Base+graph+CL}, along with their corresponding reference impressions, are randomly selected for expert evaluation \cite{attend}.
Besides, we follow \newcite{hu2021word} to utilize four metrics: Key, Readability, Accuracy, and Completeness, respectively.
We invite three medical experts to score these generated impressions based on these four metrics, with the results shown in Figure \ref{fig:human_evaluation}.
\textcolor{black}{
On the one hand, compared to \textsc{Base}, we can find that our model outperforms it on all four metrics, where 16\%, 25\%, 18\%, and 8\% of impressions from our model obtain higher quality than \textsc{Base}.
On the other hand, comparing our model against reference impressions, our model obtains close results on key, accuracy, and completeness, with 86\%, 78\%, and 92\% of our model outputs being at least as good as radiologists, while our model is less preferred for readability with a 10\% gap.
The main reason might be that many words removed in positive examples are used to keep sequence fluently, and our model tends to identify them as secondary information, leading that our model obtains relatively worse results on the readability metric.
}
}

\subsection{Analyses}
We conduct further analyses on Findings Length and Case Study.

\begin{figure}[t]
\centering
\includegraphics[width=0.5\textwidth, trim=0 0 0 10]{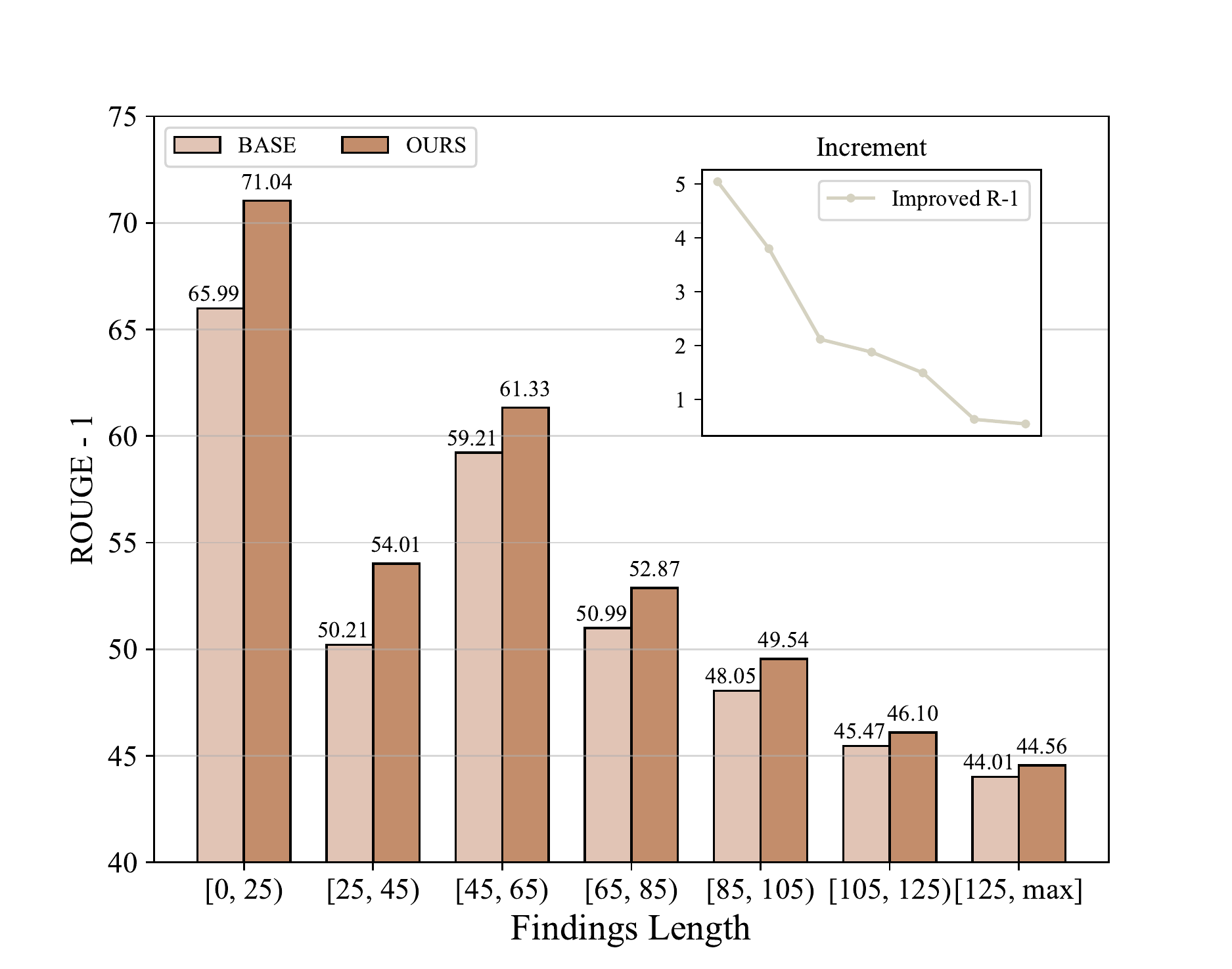}
\vskip -0.5em
\caption{R-1 score of generated impressions from \textsc{Base} and our model on the MIMIC-CXR test set, where OURS represent the \textsc{Base+graph+CL}.}
\label{fig:findings-length}
\vskip -1em
\end{figure}

\paragraph{Findings Length}
\textcolor{black}{
To test the effectiveness of the word-based length of findings, we categorize the findings on the MIMIC-CXR test set into seven groups and present the R-1 score for each group in Figure \ref{fig:findings-length}.
We have the following observations.
First, as the findings length becomes long, the performance of \textsc{Base} and our model tend to decrease, except for the second group, i.e., [25, 45], since short text are more accessible for the encoder to capture valid features, which is consistent with previous studies \cite{dai2019transformer}.
Second, our model outperforms \textsc{Base} in all the groups, further illustrating the effectiveness of our model regardless of the findings length.
Third, we can observe a grey line with a downward trend from the incremental chart in the upper right corner of Figure \ref{fig:findings-length}, indicating that our model (i.e., \textsc{Base+graph+CL}) tends to gain better improvements over \textsc{Base} on shorter findings than that on longer ones.
This is because longer findings usually contain relatively more secondary information such that it is more challenging for contrastive learning to distinguish critical knowledge.
}
\begin{figure*}[t]
\centering
\includegraphics[width=0.98\textwidth, trim=0 0 0 0]{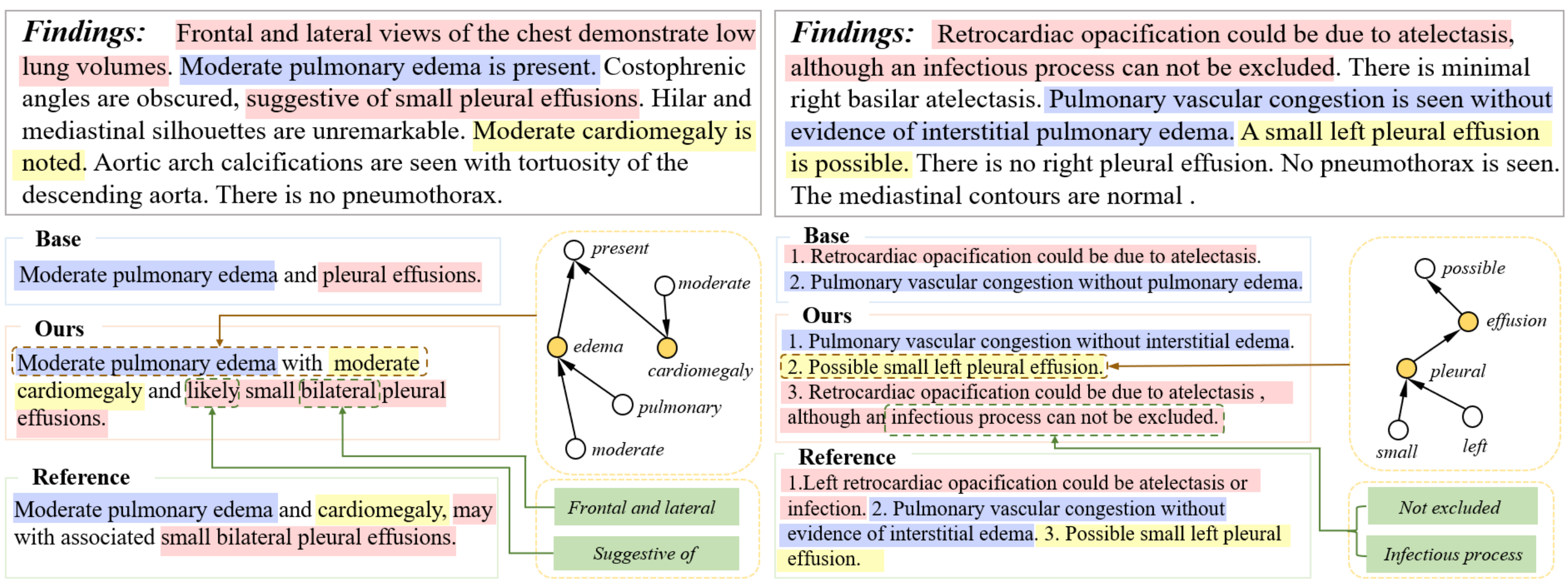}
\vskip -0.5em
\caption{Examples of the generated impressions from \textsc{Base} and \textsc{Base+graph+CL} as well as reference impressions. The yellow nodes in the graph indicate that these words are contained in entities.}
\label{fig:case}
\vskip -1em
\end{figure*}

\paragraph{Case study}
\textcolor{black}{
To further demonstrate how our approach with graph and contrastive learning helps the generation of findings, we perform qualitative analysis on two cases, and the results are shown in Figure \ref{fig:case}, where different colors on the texts indicate different critical information.
Compared to \textsc{Base} model, our model can generate more complete impressions which cover almost all the crucial abnormalities. In contrast, the \textsc {Base} model fails to identify all the key information, e.g., (\textit{``moderate cardiomegaly''} in the left example and \textit{``possible small left pleural effusion''} in the right case).
Besides, our model can generate more accurate impressions with an appropriate word to represent possibility and a better modifier to describe the observation.
On the one hand, in Figure \ref{fig:case}, \textit{``suggestive of''} in the left example and \textit{``may''} in the right example imply a type of uncertainty, which means that doctors wonder whether the abnormal observation exists when writing findings, so that the corresponding word (i.e., \textit{``likely''}) is used to describe this sensitive information.
On the other hand, in the left case, according to the phrase \textit{``Frontal and lateral''} in its original findings, our model can generate the synonym \textit{``bilateral''} to depict the symptom \textit{``pleural effusions''} more specifically.
}

\section{Related Work}
Recently, NLP technology has broadly applied in the medical domain, such as medical entity recognition \cite{liu2021exploring,zhao2019neural}, radiology report generation \cite{chen2021cross,zhang2020radiology,liu2021medical}, AIG, etc.
Impression generation can be regarded as a type of summarization task that has drawn substantial attention in recent years, and there are many studies for addressing general abstractive summarization \cite{see2017get,li2020keywords,you2019improving,huang2020knowledge}.
\newcite{you2019improving} designed a novel focus-attention mechanism and saliency-selection network, equipped in the encoder and decoder to enhance summary generation.
\newcite{li2020keywords} proposed an abstractive sentence summarization method guided by the key words, which utilized a dual-attention and a dual-copy mechanism to integrate the semantics of both original sequence and key words.
Many methods propose to introduce specific designs on the general summarization model to address radiology impression generation \cite{zhang2018learning,attend,macavaney2019ontology,hu2021word,abacha2021overview}.
\newcite{macavaney2019ontology,attend} extracted the salient clinical ontology terms from findings and then incorporated them into the summarizer through a separate encoder for enhancing AIG.
\newcite{hu2021word} further introduced pre-defined word graphs to record salient words as well as their internal relation and then employed two separate graph encoders to leverage graphs for guiding the decoding process.
Most of these approaches exploit separate encoders to encode pre-defined knowledge (e.g., ontology terms and word graph), which are then utilized to enhance impression generation.
However, they tend to over-rely on the quality of pre-extracted ontologies and word graphs and lack sensitivity to vital information of findings themselves.
Compared to these models, our method offers an alternative solution to robustly improve key information extraction with the help of both graphs and contrastive learning.

\section{Conclusion}

\textcolor{black}{
In this paper, we propose to combine graphs and contrastive learning to better incorporate valuable features for promoting impression generation.
Specifically, we utilize the graph encoder to extract relation information from the graph, constructed by medical entities and the dependence tree, for enhancing the representation from the pre-trained text encoder.
In addition, we employ contrastive learning to assist the model in distinguishing between critical and secondary information, simultaneously improving sensitivity to important word representation by comparing positive and negative examples.
Furthermore, we conduct experiments on two benchmark datasets, and the results illustrate the effectiveness of our proposed method, where new state-of-the-art results are achieved.
}

\section*{Acknowledgements}
This work is supported by Chinese Key-Area Research and Development Program of Guangdong Province (2020B0101350001), NSFC under the project “The Essential Algorithms and Technologies for Standardized Analytics of Clinical Texts” (12026610) and the Guangdong Provincial Key Laboratory of Big Data Computing, The Chinese University of Hong Kong, Shenzhen.

\bibliography{anthology,custom}
\bibliographystyle{acl_natbib}

\clearpage
\appendix
\section{Appendix}
\label{sec:appendix}

\subsection{Hyper-parameter Settings}
\label{appendix:hyperparameter}

Table \ref{Tab:hyperparameters} reports the hyper-parameters tested in tuning our models on \textsc{MIMIC-CXR} and \textsc{OpenI}.
For each dataset, we try all combinations of the hyper-parameters and use the one achieving the highest R-1 for \textsc{MIMIC-CXR} and \textsc{OpenI}.

\begin{table}[t]
\footnotesize
\centering
\resizebox{.49\textwidth}{!}{
\begin{tabular}{l|l|l}
\toprule[1pt]
{\textsc{\textbf{Model}}}&\textsc{\textbf{Hyper-Parameter}}  &\textsc{\textbf{Value}} \\
\hline                       

\multirow{3}{*} {\makecell*[l]{\textsc{MIMIC-CXR}}}
& \textsc{Batch Size} & {32,64,\textbf{128},300} \\
& \textsc{Learning Rate} & {8e-5,\textbf{2e-4}, 1e-3, 0.05,}  \\
& \textsc{Training steps} & {150000}  \\

\midrule

\multirow{3}{*} {\makecell*[l]{\textsc{OpenI}}}
& \textsc{Batch Size} & {32,64,\textbf{128},300} \\
& \textsc{Learning Rate} & {8e-5,\textbf{5e-3}, 1e-3, 0.05}  \\
& \textsc{Training steps} & {20000}  \\
\bottomrule
 \end{tabular}}
 \linespread{1}
  \caption{The hyper-parameters that we have experimented on the datasets. The bold values illustrates the best configurations of different models.}%
  \label{Tab:hyperparameters}
\end{table}

\end{document}